\documentclass[conference]{IEEEtran}

\usepackage{makecell}
\usepackage{adjustbox}
\usepackage{diagbox}
\usepackage{mathtools} 
\usepackage{amsfonts}
\usepackage{amsmath}
\usepackage{mathrsfs}
\usepackage{balance}
\usepackage{soul,color}
\usepackage{pifont}
\usepackage{dsfont}

\usepackage{cite}
\usepackage{graphicx}
\usepackage{textcomp}
\usepackage[table,xcdraw]{xcolor}
\usepackage{ragged2e}
\usepackage{bbm}
\usepackage{multirow}
\usepackage{algpseudocode}
\usepackage[caption=false, font=footnotesize]{subfig}
\usepackage[hang,flushmargin]{footmisc}
\usepackage{lipsum}
\usepackage[flushleft]{threeparttable}
\usepackage{etoolbox}

\usepackage{array}
\newcolumntype{C}[1]{>{\centering\let\newline\\\arraybackslash\hspace{0pt}}m{#1}}
\def\BibTeX{{\rm B\kern-.05em{\sc i\kern-.025em b}\kern-.08em
    T\kern-.1667em\lower.7ex\hbox{E}\kern-.125emX}}
    
\makeatletter

\usepackage{hyperref}
\hypersetup{
    colorlinks=true,
    linkcolor=magenta,
    filecolor=blue,      
    urlcolor=cyan,
    citecolor=green,
}

\usepackage{tabularx}
\newcolumntype{Y}{>{\centering\arraybackslash}X}

\begin{document}

\newcommand{\MYfooter}{\smash{
\hfil\parbox[t][\height][t]{\textwidth}{\centering
\thepage}\hfil\hbox{}}}

\makeatletter

\makeatother
\pagestyle{headings}
\addtolength{\footskip}{0\baselineskip}
\addtolength{\textheight}{0\baselineskip}

\title{Monkeypox Skin Lesion Detection Using Deep Learning Models: A Feasibility Study}

\author{
\IEEEauthorblockN{Shams Nafisa Ali\textsuperscript{1}, Md. Tazuddin Ahmed\textsuperscript{1}, Joydip Paul\textsuperscript{1}, Tasnim Jahan\textsuperscript{1}, S. M. Sakeef Sani\textsuperscript{1}, \\
Nawsabah Noor\textsuperscript{2}, Taufiq Hasan\textsuperscript{1}, \emph{Senior Member, IEEE}}
\IEEEauthorblockA{\text{\textsuperscript{1}mHealth Lab, Department of Biomedical Engineering, Bangladesh University of Engineering and Technology} \\
\text{\textsuperscript{2}Popular Medical College, Dhaka, Bangladesh.}
Email: taufiq@bme.buet.ac.bd}}
\maketitle
\begin{abstract}
The recent monkeypox outbreak has become a public health concern due to its rapid spread in more than 40 countries outside Africa. Clinical diagnosis of monkeypox in an early stage is challenging due to its similarity with chickenpox and measles. In cases where the confirmatory Polymerase Chain Reaction (PCR) tests are not readily available, computer-assisted detection of monkeypox lesions could be beneficial for surveillance and rapid identification of suspected cases. Deep learning methods have been found effective in the automated detection of skin lesions, provided that sufficient training examples are available. However, as of now, such datasets are not available for the monkeypox disease. In the current study, we first develop the ``Monkeypox Skin Lesion Dataset (MSLD)" consisting skin lesion images of monkeypox, chickenpox, and measles. The images are mainly collected from websites, news portals, and publicly accessible case reports. Data augmentation is used to increase the sample size, and a 3-fold cross-validation experiment is set up. In the next step, several pre-trained deep learning models, namely, VGG-16, ResNet50, and InceptionV3 are employed to classify monkeypox and other diseases. An ensemble of the three models is also developed. ResNet50 achieves the best overall accuracy of $82.96(\pm4.57\%)$, while VGG16 and the ensemble system achieved accuracies of $81.48(\pm6.87\%)$ and $79.26(\pm1.05\%)$, respectively. 
A prototype web-application is also developed as an online monkeypox screening tool.  
While the initial results on this limited dataset are promising, a larger demographically diverse dataset is required to further enhance the generalizability of these models.
\end{abstract}

\begin{IEEEkeywords}
\textit{Computer-aided diagnosis, Skin lesion detection, Monkeypox, Deep learning.}
\end{IEEEkeywords}

\section{Introduction}
As the world recovers from the COVID-19 pandemic, the recent multi-country outbreak of monkeypox has raised concerns in global communities.
The World Health Organization (WHO) declared that the outbreak poses a moderate risk to global public health and has stopped short of declaring it a public health emergency.
However, healthcare organizations such as World Health Network (WHN) expressed a heightened concern~\cite{WHN} and highlighted the need for immediate and concerted global action against the disease.
\par
Monkeypox is a zoonotic disease from the genus \textit{Orthopoxvirus}. It closely resembles chickenpox, measles, and smallpox regarding clinical features~\cite{facts}. The minor differences in the skin rash of these diseases, coupled with the relative rarity of monkeypox have made the early diagnosis of this condition very challenging for healthcare professionals. On the contrary, the confirmatory PCR test is also not widely available. Although the case fatality ratio has been reported to be 3–6\% for the recent outbreak~\cite{facts}, early detection of monkeypox, corresponding contact tracing, and isolation are essential to limit the community transmission of the virus. In this scenario, AI-based automated computer-aided systems may substantially limit its global spread.
\par 
\begin{figure}[t]
    \center
    \includegraphics[width=\linewidth]{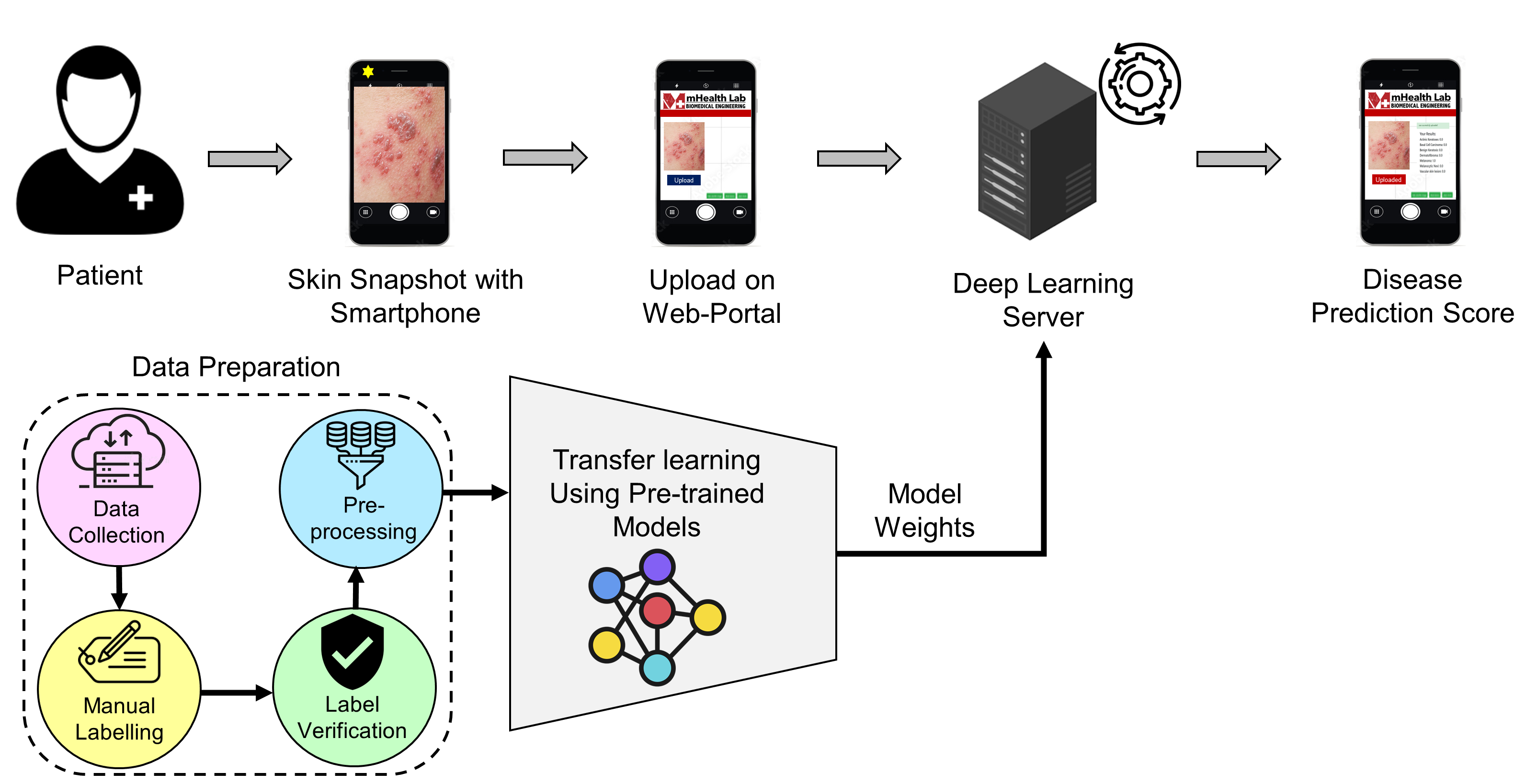}
    \caption{A flow-diagram of the proposed monkeypox detection system. A prototype web-app is developed that incorporates the developed deep learning models to detect monkeypox from skin lesions uploaded by users.}
    \label{GA}
\end{figure}

In recent years, the multi-faceted applications of deep learning (DL), particularly the variations of Convolutional Neural Networks (CNNs), have revolutionized different domains of medical science due to their superior learning capability~\cite{ravi2016deep, wang2019deep}. When trained with a large number of data, these deep networks can process images in different layers, automatically extracting salient features and learning to identify the optimal representations for specified tasks~\cite{hosny2019classification}. However, the requirement for large amounts of data and time-consuming training with dedicated computational resources limits the applicability of DL-based frameworks~\cite{TL}. While using accelerators (e.g., GPU, TPU) resolves the time and resource-related issues, the dataset-related concerns persist due to the difficulty of obtaining unbiased, homogeneous medical data. Data augmentation~\cite{aug} is a well-known method of increasing the dataset size by generating additional samples through slight modifications of existing data. In case of scarcity of data, transfer learning~\cite{TL} is also a commonly used technique. This method utilizes a CNN model pre-trained on a large dataset (e.g., ImageNet) 
and transfers its knowledge for context-specific learning on a different, comparatively smaller dataset.


\par
Currently, there is no publicly available monkeypox skin lesion dataset for developing automated detection algorithms. There are impediments considering privacy and validity concerns. Moreover, the high prevalence of monkeypox in the under-developed African regions may introduce a bias in the dataset because of very high inter-class similarity and intra-class variability. In this paper, we first introduce the ``Monkeypox Skin Lesion Dataset (MSLD)"\footnote{\url{https://github.com/mHealthBuet/Monkeypox-Skin-Lesion-Dataset}},
an openly accessible dataset containing web-scrapped images of different body parts (face, neck, hand, arm, leg) of patients with monkeypox and non-monkeypox (measles, chickenpox) cases. We also present a DL-based preliminary feasibility study leveraging transfer learning involving VGG16~\cite{VGG}, ResNet50~\cite{resnet} and InceptionV3~\cite{inception} architectures to explore the potential of deep learning models for the early detection of the monkeypox disease. Additionally, we have soft-launched a web-app\footnote{\url{https://monkey-pox-detector-mhealthlab.herokuapp.com/}} built on the open-source streamlit framework that is capable of analysing the uploaded image and predicting whether a subject is a potential monkeypox suspect and he/she should consult a physician on an urgent basis or not. Our current working pipeline is graphically represented in Fig.~\ref{GA}.


\begin{figure}[t]
    \center
    \includegraphics[width=0.8\linewidth]{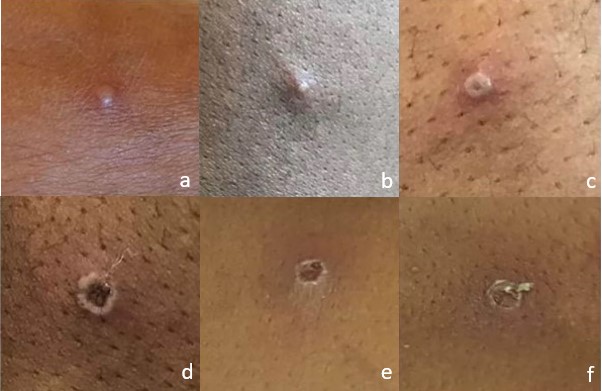}
    \caption{The Monkeypox lesion through its various stages: (a) early vesicle, (b) small pustule ($\oslash~2$mm), (c) umbilicated pustule ($\oslash~3$-$4$mm), (d) ulcerated lesion ($\oslash~5$mm), (e) crusting of a mature lesion, (f) partially removed scab.}
    \label{stages}
  \end{figure}


\section{Background} 
Monkeypox virus, an enveloped double-stranded DNA virus, is one of the members of the \textit{Poxviridae} family, which is closely related to the variola virus~\cite{facts}. Various animal species, including squirrels, Gambian pouched rats, dormice, and non-human primates, were first identified as the natural hosts of monkeypox virus~\cite{facts}. The first human case of monkeypox was confirmed in the Democratic Republic of Congo (DRC) in 1970, and subsequently, it has been endemic in the tropical rainforests of west and central Africa~\cite{facts}. 
\par 
Since the beginning of 2022, monkeypox outbreaks have been reported in diverse countries across the globe. Currently, 5135 laboratory-confirmed cases have been reported from 66 member states, including the Regions of America, Europe, Africa, Eastern Mediterranean, and Western Pacific (updated 30 June 2022)~\cite{monkeymeter}. After the rapid spread of the disease in non-endemic countries with no epidemiological connections to endemic areas, the disease was declared a moderate global health risk as assessed by the WHO~\cite{facts}.
\par 
Monkeypox is usually a self-limited disease; the symptoms last 2 to 4 weeks. Children are mostly affected severely. The severity of the disease depends on the extent of virus exposure, the health status of the patient, and the complications. The virus's incubation period is in the range of 5 to 21 days~\cite{facts}. The invasion period (0-5 days) is characterized by the symptoms of fever, lymphadenopathy (swollen lymph nodes), myalgia (muscle ache), asthenia (physical weakness), and severe headache. The rash begins within 1-3 days of fever onset and is usually noticed on the face, palms of the hands, and soles of the feet. However, it can also be found on the mouth, eyes, and genitals~\cite{facts}. In the skin eruption phase (2-4 weeks), the lesions follow the four-stage progression; macules (lesions with a flat base) to papules (raised firm painful lesions) to vesicles (filled with clear fluid) to pustules (filled with pus) followed by encrustation.
\par 
The gold standard diagnosis of monkeypox is done through 
histopathology and virus isolation. In addition, a PCR test can help for the confirmation of the diagnosis \cite{Monkeypox_CDC}. However, when these diagnostic tools are not readily available, early diagnosis based on clinical examination of the skin lesions can be beneficial. With the ubiquitous availability of smartphones, AI-based skin lesion detection systems can assist in diagnosis and bridge the gap in the healthcare systems.

\begin{table}[t]
\centering
\caption{Distribution of the Monkeypox Skin Lesion Dataset (MSLD)}
\footnotesize
\label{data}
\scalebox{1}{\begin{tabular}{c|c|c|c} 
\hline
\hline
\bf Class label & \begin{tabular}[c]{@{}c@{}} \bf No. of Original \\  \bf Images\end{tabular} & \begin{tabular}[c]{@{}c@{}} \bf No. of Unique \\  \bf Patients\end{tabular} & \begin{tabular}[c]{@{}c@{}} \bf No. of Augmented \\  \bf Images\end{tabular} \\
\hline
Monkeypox & 102               &        55           & 1428               \\
Others     & 126               &      107             & 1764               \\\hline
Total      & 228               &        162           & 3192         \\     
\hline\hline
\end{tabular}}
\end{table}

\section{Dataset Preparation}
Our monkeypox skin lesion dataset is primarily compiled from publicly available case reports, news portals, and websites through extensive manual searching. We did not use automatic web-scrappers. 
In this work, we mainly focus on distinguishing the `Monkeypox' cases from similar non-monkeypox cases~\cite{facts}. Thus, we include skin lesion images of chickenpox and measles as the `Others' class and prepare the dataset for binary classification

\par 
All the skin lesion images were verified using Google's Reverse Image Search and cross-referenced with other sources. 
Through a 2-stage screening process, the out-of-focus, low-resolution, and low-quality images were discarded, and only the unique images that satisfy the quality requirements were selected. Next, the images were cropped to their region of interest and resized to $224\times224$ pixels while maintaining the aspect ratio.
\par 
\begin{figure}[!ht] 
    \center
    \includegraphics[width=\columnwidth]{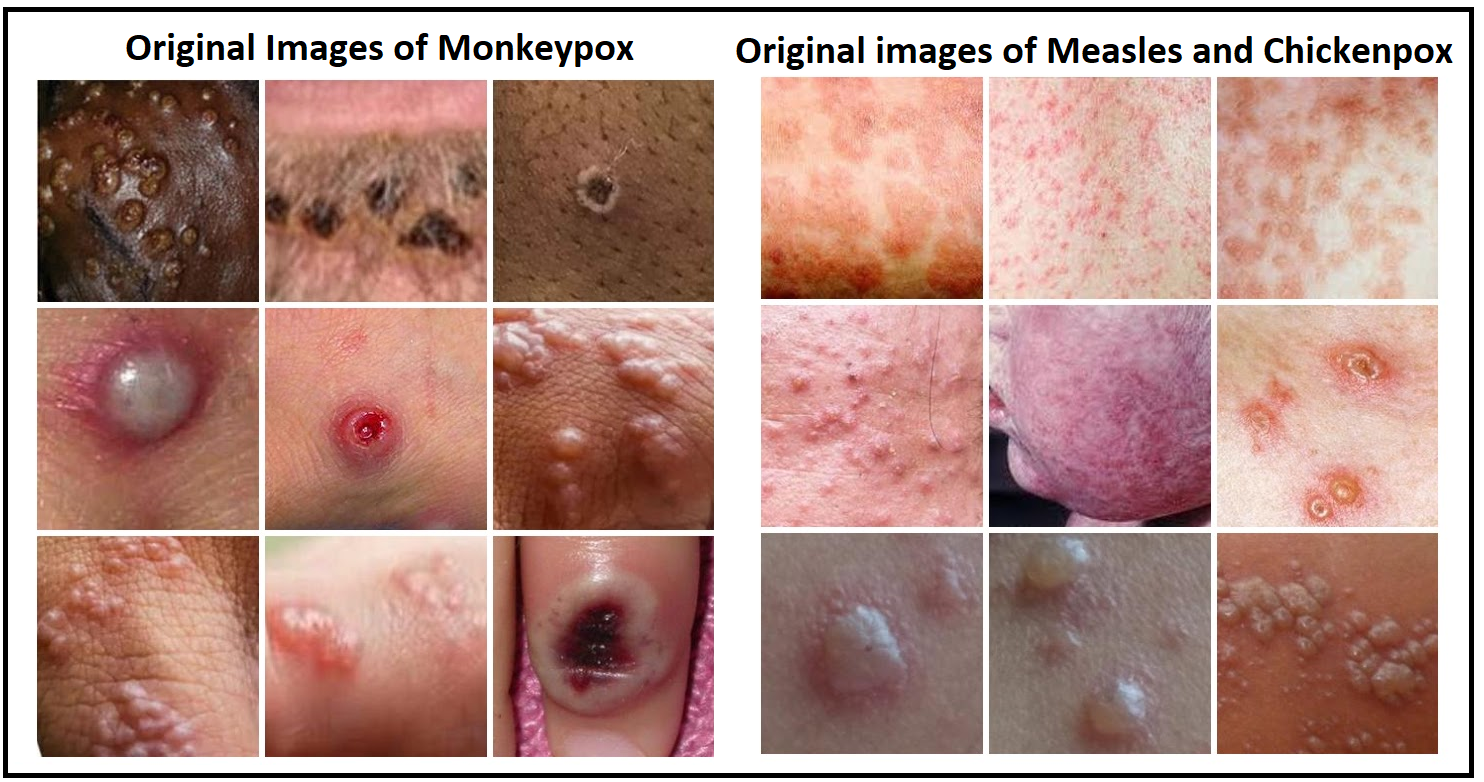}
    \caption{Sample images from the collected dataset. Images from the `Monkeypox' class are shown on the left panel, while the `Others' class (measles and chickenpox) images are shown on the right.}
    \label{original_image}
\end{figure}

We have collected a total number of $228$ images, among which $102$ belongs to the `Monkeypox' class and the remaining $126$ represents the `Others' class (i.e., chickenpox and measles). A few representative samples are shown in Fig.~\ref{original_image}. Since this is the early phase of the monkeypox outbreak, reliable data are scarce. Hence, to aid the classification task, several data augmentation methods, including rotation, translation, reflection, shear, hue, saturation, contrast and brightness jitter, noise, and scaling, have been applied to increase the dataset size. 
Post-augmentation, the number of images increased by approximately 14-folds, with the classes `Monkeypox' and `Others' consisting of $1428$ and $1764$  images, respectively. The augmented images are also provided in a separate folder in the dataset to ensure reproducibility. However, based on their requirement, the users may also apply different types of image augmentor on the original images. A detailed distribution of the dataset is provided in Table~\ref{data}.

\section{Experiments}
\subsection{Pre-trained Models}
In this work, we select three well-known CNN architectures namely, VGG16~\cite{VGG}, ResNet50~\cite{resnet} and InceptionV3~\cite{inception}, pre-trained on the ImageNet dataset. These models are chosen as they have demonstrated excellent classification performance across different fields of computer vision, including medical images through transfer learning. 
\par 
\begin{figure}[ht]
    \center
    \includegraphics[width=0.8\columnwidth]{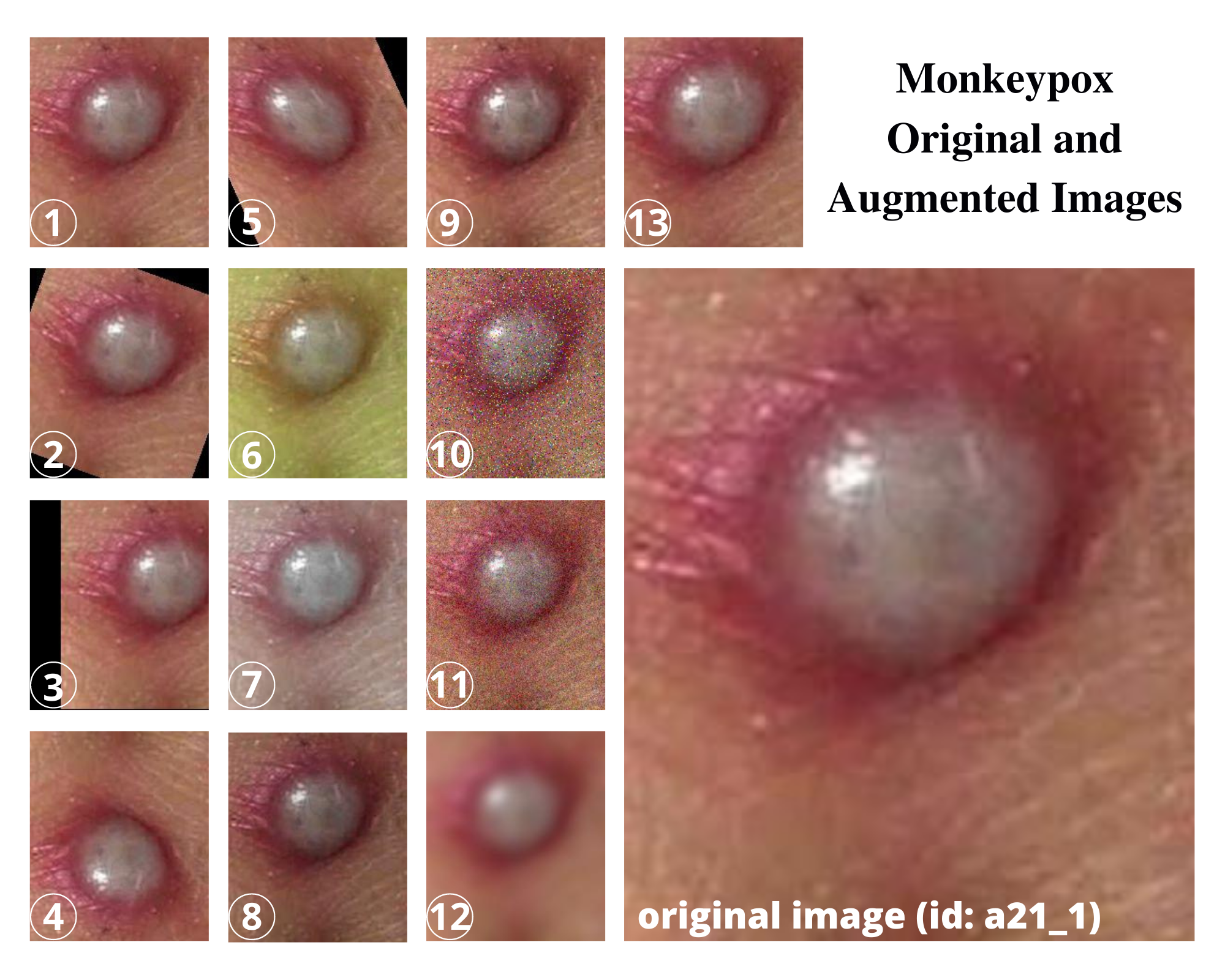}
    \caption{An original monkeypox skin lesion image and its 13 augmented versions used for training. These created as follows: (1) Random rotation by a multiple of 90 degrees, (2) Random rotation in the range of -45 to 45 degrees, (3) Translation, (4) Reflection, (5) Shear, (6) Hue jitter, (7) Saturation jitter, (8) Brightness jitter, (9) Contrast jitter, (10) Salt and Pepper Noise, (11) Gaussian Noise, (12) Synthetic Blur and (13) Scaling.}
    \label{aug_image}
\end{figure}
VGG16 is one of the first networks to investigate CNN's depth to improve accuracy in large-scale image recognition tasks~\cite{VGG}. It consists of $3\times3$ convolutional filters applied in several layers. The concept of utilizing Factorized Convolution enabled feature extraction from increased depth but simultaneously ensured that the model did not overfit the training data. The ResNet model was formulated based on the concept of residual learning~\cite{resnet}. It consists of residual modules where convolution operations are followed by Batch Normalization and ReLU non-linearity. These blocks enable the inputs to forward propagate faster and extract features more efficiently. Because of the “stacked” Inception modules, despite having fewer parameters than VGG16, the Inception architecture can perform better back-propagation~\cite{inception}. Since its first release, several changes have been made to the original architecture, and improved versions have been released.

\subsection{Implementation Details}
Input images with dimensions $(224, 224, 3)$ were fed into the selected pre-trained models. The fully connected layers were removed. Although the architectures have varying depths, we experimented with keeping different numbers (4/6/8/12) of bottom layers trainable to ensure homogeneity and better generalization. After extensive experimentation, we decided to unfreeze the bottom eight layers. Next, we flatten the backbone model’s output, followed by three blocks of fully connected (FC) layers, and dropout to the network. The FC layers had successively $4096$, $1072$ and $256$ nodes while the dropout factors were respectively $0.3$, $0.2$ and $0.15$. Finally, an FC layer with two nodes was employed with a softmax activation function for this binary classification task.
\par 
The network architectures were implemented in Keras and were accelerated using Nvidia K80 GPUs provided by Kaggle notebooks. The batch size was set to 16. The adaptive learning rate optimizer (Adam) with an initial learning rate of $10^{-5}$ and binary cross-entropy loss function was employed for training. 
\par 


\begin{figure}[b]
    \center
    \includegraphics[width=\columnwidth]{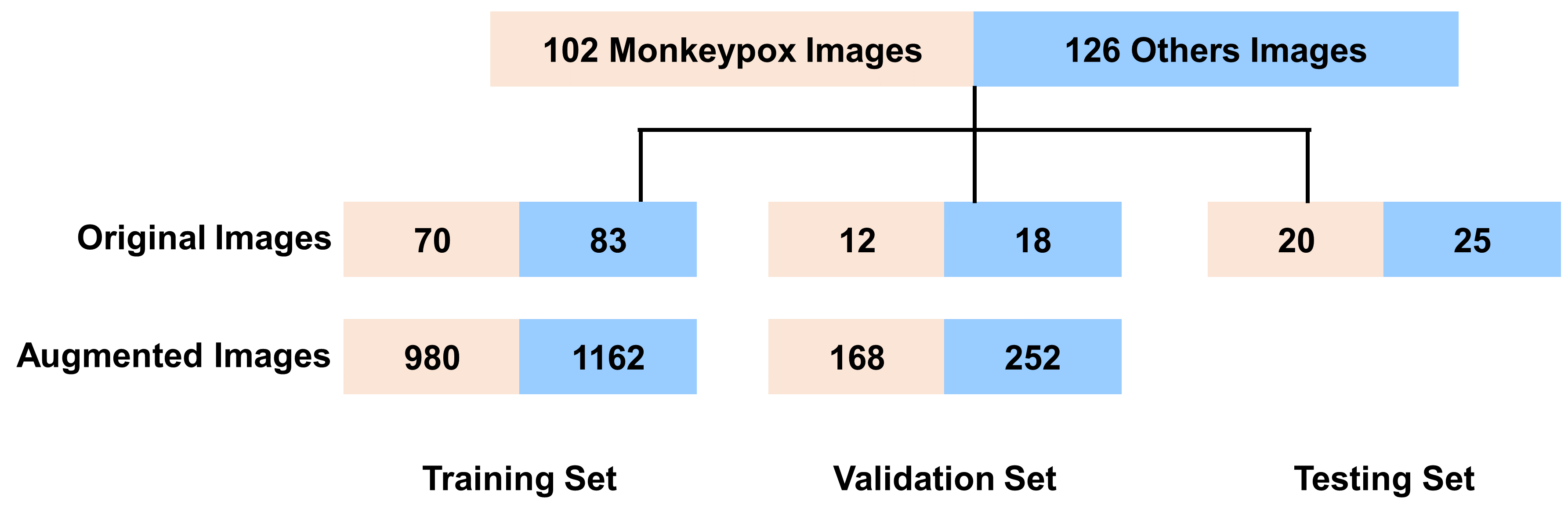}
    \caption{An overview of the train-test split used for 3-fold cross-validation. Data augmentation is used for the training and validation set only.}
    \label{Dataset splits}
\end{figure}

\subsection{Experimental Design}
A three-fold cross-validation experiment was conducted. The original images were split into training, validation, and test set(s) with the approximate proportion of 70:10:20 while maintaining patient independence. According to the commonly perceived data preparation practice, the training and validation images were augmented while the test set contained only the original images (see Fig.~\ref{Dataset splits}). Accuracy, precision, F1-score, sensitivity are used as performance metrics. 




\subsection{Results}
We evaluated the performance of the selected pre-trained models for the 3-fold cross-validation experiment. The obtained results are summarized in Table~\ref{t3}. 
While ResNet50 yields the best accuracy ($82.96\pm 4.57\%$), VGG16 shows competitive performance ($81.48\pm6.87\%$). We also implement an ensemble of the 3 models by employing majority voting. Based on averaged performance across the 3-folds, the ensemble did not show superior results compared to the best-performing ResNet50 model. However, the ensemble system shows the lowest standard deviation of the accuracy metric, indicating that its performance is more consistent across the 3 folds.
\par
\begin{figure}[t]
    \center
    \includegraphics[width=0.9\linewidth]{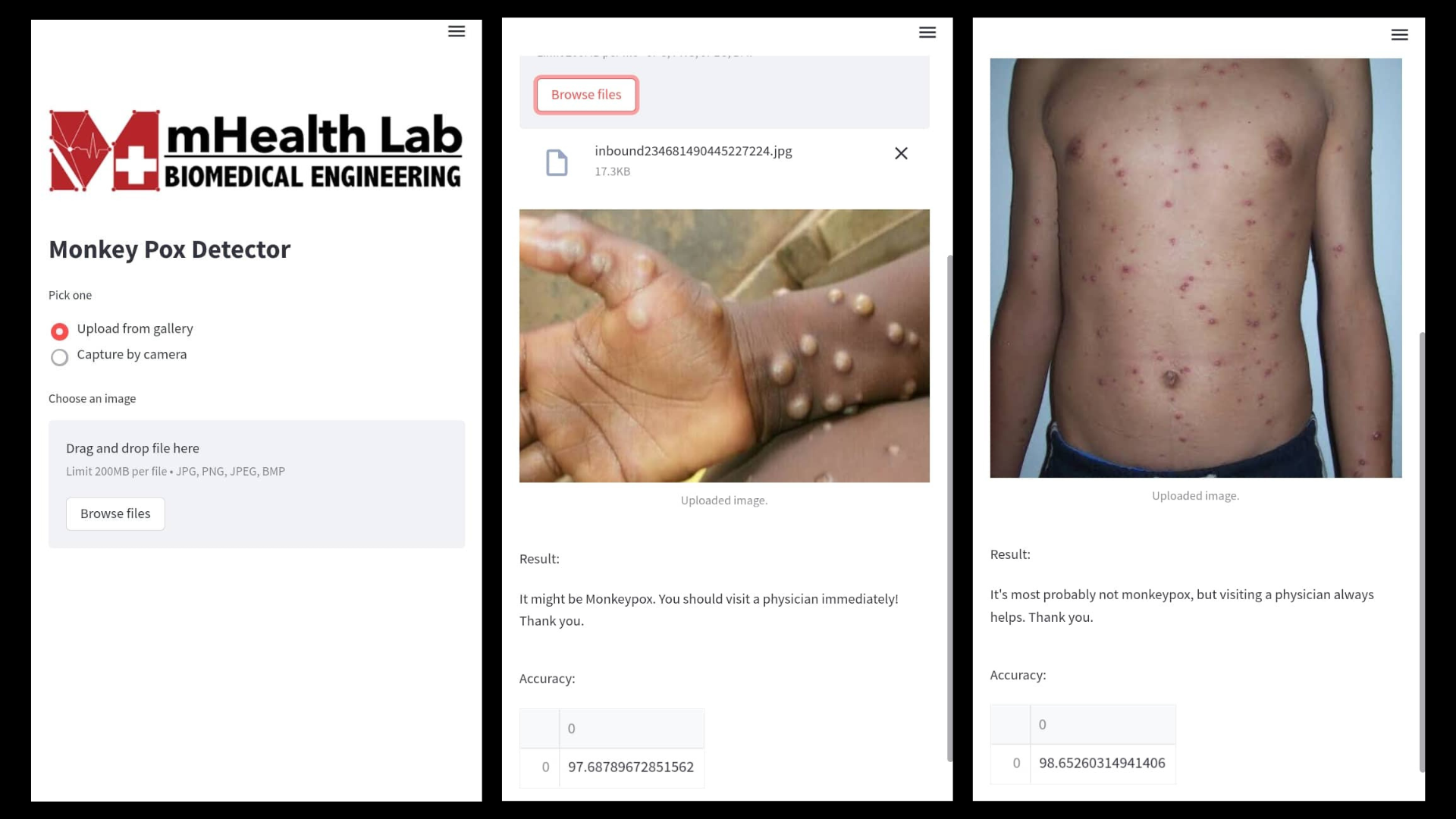}
    \caption{The interface of the online monkeypox screening tool.}
    \label{app interface}
\end{figure}

 
The best performing model has been deployed in a prototype web-app. Users can upload a photo of their skin lesions and receive an initial assessment for monkeypox (see Fig.~\ref{app interface}).

\begin{table}[hb]
\centering
\caption{Performance Comparison of Different Deep Learning Models and The Proposed Ensemble}
\label{t3}
\scalebox{0.9}{\begin{tabular}{ c | c| c |c | c  } 
\hline
\textbf{Network} & {\textbf{Accuracy (\%)}} & {\textbf{Precision}} & {\textbf{Recall}} &{\textbf{F1 score}}\\
\hline
\hline
VGG16 & 81.48$ \pm$ 6.87  & 0.85 $\pm$ 0.08  & 0.81 $\pm$ 0.05 & 0.83 $\pm$ 0.06 \\ 
ResNet50 & 82.96$ \pm$ 4.57 & 0.87 $\pm$ 0.07 & 0.83 $\pm$ 0.02 &  0.84 $\pm$ 0.03  \\ 
InceptionV3 & 74.07$ \pm$ 3.78 & 0.74 $\pm$ 0.02 & 0.81 $\pm$ 0.07 & 0.78 $\pm$ 0.04 \\ 
Ensemble & 79.26$ \pm$ 1.05 & 0.84 $\pm$ 0.05 & 0.79 $\pm$ 0.07 & 0.81 $\pm$ 0.02 \\ 
\hline
\end{tabular}}
\end{table}



\section{Discussion}
In this work, we present our preliminary findings of monkeypox skin lesion detection using well-known deep learning architectures. Although the classification performance obtained is quite promising, certain constraints limit the wider applicability of the results. Firstly, the current number of unique patients in the dataset is limited. This reduces the generalization capability of these models. If more samples can be included with a better geographical, racial, and gender distribution, it is possible to achieve a consistent performance across such variabilities. Secondly, as a initial approach, we have used the pre-trained weights from the ImageNet dataset for transfer learning which does not contain any skin lesion images. Thus, the accuracy and generalizability of the system could be further improved by using a multi-source dermatoscopic image dataset for pre-training the models. 
Furthermore, the dataset is created primarily by web-scraping, which lacks relevant meta-data that is vital for diagnosis, e.g., patient's clinical history, number of days since disease onset, and the disease stage. A more consorted effort and international collaboration is needed to collect a larger dataset that can provide generalizable results across different demographics.



\section{Conclusion}
In this study, we have presented the open-source ``Monkeypox Skin Lesion Dataset (MSLD)" for automatic detection of monkeypox from skin lesions and performed an initial feasibility study using state-of-the-art deep learning architectures (VGG16, ResNet50, InceptionV3) leveraging the transfer learning approach. Despite being a small dataset, the promising results obtained after 3-fold cross-validation reveal the potential to use AI-assisted early diagnosis of this disease.
We are hopeful that this dataset will create new avenues for the researchers in terms of developing remotely deployable computer-aided diagnostic tools for wide-scale screening and early detection of monkeypox, especially in cases where traditional testing methods are unavailable. We also believe that our soft-launched web-app prototype will  assist the monkeypox suspects to conduct preliminary screening from the comforts of home and enable them to take adequate action in the early stages of the infection. 


\bibliographystyle{IEEEtran}
\bibliography{ref}

\end{document}